\documentclass[letterpaper, 10 pt, conference]{ieeeconf}  

\IEEEoverridecommandlockouts                              

\usepackage{amsmath}
\usepackage{graphicx}
\usepackage{booktabs}
\usepackage{multirow}
\usepackage {color}
\usepackage{amssymb}
\usepackage[table]{xcolor}
\usepackage{makecell}

\title{\LARGE \bf
3D-JEPA: A Joint Embedding Predictive Architecture for 3D Self-Supervised Representation Learning
}

\author{Naiwen Hu, Haozhe Cheng, Yifan Xie, Shiqi Li and Jihua Zhu$^{*}$
\thanks{$^{*}$:corresponding author (zhujh@xjtu.edu.cn). The authors are with School of 
Software Engineering, Xi'an Jiaotong University, Xi'an710000,
China and Shaanxi Joint Key Laboratory for Artifact Intelligence, China.}
}

\begin{document}

\maketitle
\thispagestyle{empty}
\pagestyle{empty}

\begin{abstract}

Invariance-based and generative methods have shown a conspicuous performance for 3D self-supervised representation learning (SSRL). However, the former relies on hand-crafted data augmentations that introduce bias not universally applicable to all downstream tasks, and the latter indiscriminately reconstructs masked regions, resulting in irrelevant details being saved in the representation space. To solve the problem above, we introduce 3D-JEPA, a novel non-generative 3D SSRL framework. Specifically, we propose a multi-block sampling strategy that produces a sufficiently informative context block and several representative target blocks. We present the context-aware decoder to enhance the reconstruction of the target blocks. Concretely, the context information is fed to the decoder continuously, facilitating the encoder in learning semantic modeling rather than memorizing the context information related to target blocks. Overall, 3D-JEPA predicts the representation of target blocks from a context block using the encoder and context-aware decoder architecture. Various downstream tasks on different datasets demonstrate 3D-JEPA's effectiveness and efficiency, achieving higher accuracy with fewer pretraining epochs, e.g., 88.65\% accuracy on PB\_T50\_RS with 150 pretraining epochs.

\end{abstract}

\section{INTRODUCTION}

Point cloud has attracted widespread attention as the primary modality for 3D perception, including autonomous driving~\cite{liu2023bevfusion,zhou2024lidarformer}, simultaneous localization and mapping (SLAM)~\cite{duan20243d}. However, acquiring point cloud labels is time-consuming and expensive making point cloud understanding difficult. Self-supervised representation learning (SSRL)~\cite{devlin2018bert,he2022masked, roggiolani2023unsupervised,nunes2022segcontrast} aims to learn transferable representation from unlabeled data, which benefits a variety of downstream tasks through fine-tuning. Inspired by the significant improvements of SSRL methods over the supervised learning counterpart in the fields of 2D~\cite{he2022masked,he2020momentum} and Natural Language Processing (NLP) ~\cite{devlin2018bert}. Recently, Point-MAE~\cite{pang2022masked} and Pointclip~\cite{zhang2022pointclip} have achieved superior performance in 3D vision tasks. 

The mainstream SSRL methods can be divided into invariance-based methods~\cite{afham2022crosspoint} and generative methods~\cite{pang2022masked,10611402}. The former optimizes the model to produce similar embeddings for positive sample pairs~\cite{oord2018representation}. The positive-negative sample pairs are constructed during pretraining by hand-crafted point cloud data augmentations such as rotate, scale and translate~\cite{huang2021spatio}. However, it also introduces additional bias and is not applicable for all downstream tasks~\cite{assran2022hidden}. In addition, the latter masks or removes portions of the input data and reconstructs the corrupted content at the pixel or token level. Despite its effectiveness~\cite{yu2022point,pang2022masked}, most methods predict representation at the pixel level and reconstruct every bit of missing information. As a result, previous 3D generative methods produce a lower semantic level representation and focus too much on irrelevant details instead of capturing high-level predictable concepts.
   \begin{figure}
      \centering
      \includegraphics[width=0.45\textwidth]{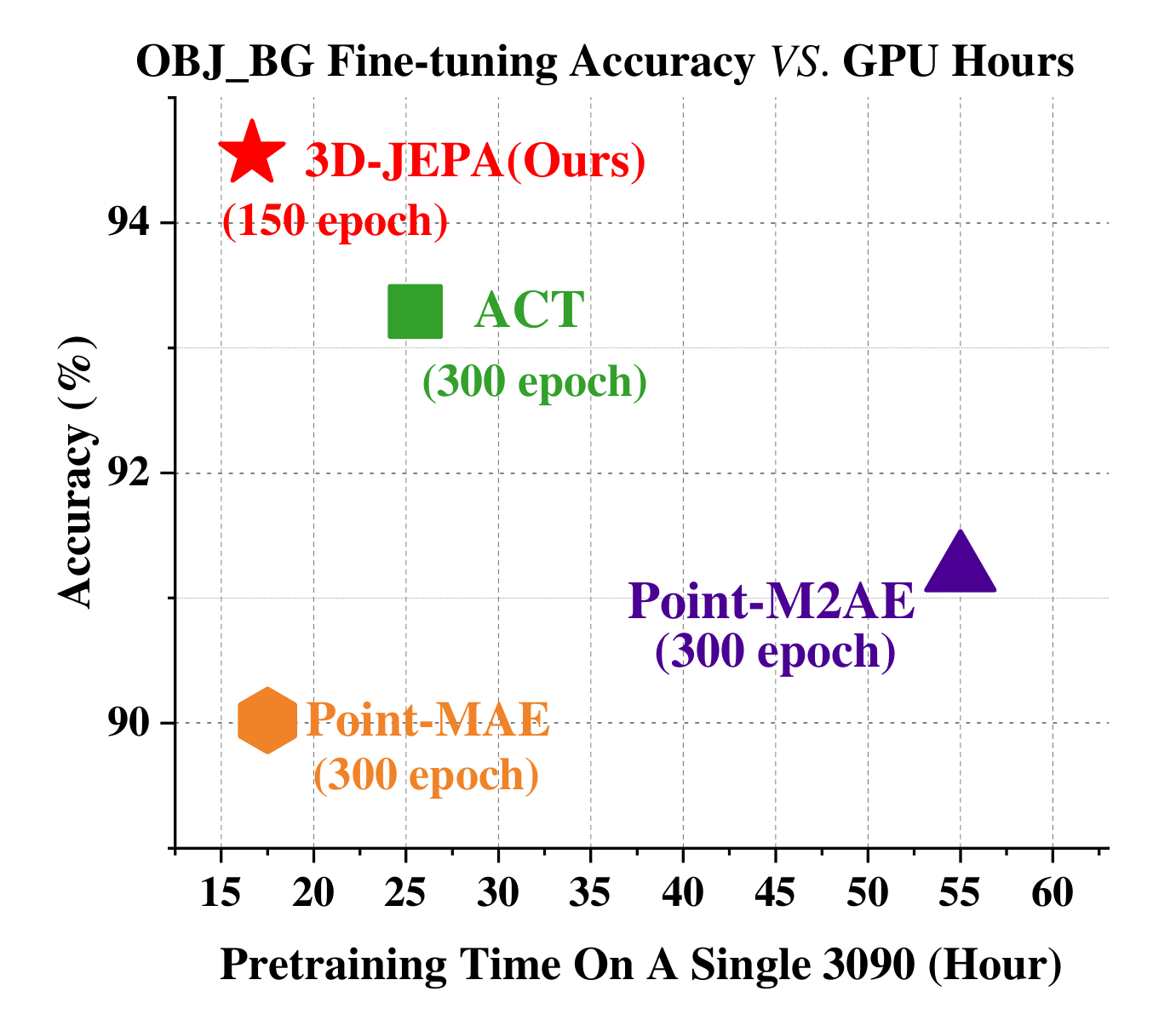}
      \vspace{-0.2cm}
      \caption{\textbf{OBJ\_BG Fine-tuing Classification Accuracy.} By predicting target blocks from a single context block without any data augmentations, the 3D-JEPA learns strong point cloud representation with less computing.}
      \label{teaser}
      \vspace{-0.5cm}
   \end{figure}

   \begin{figure*}
      \centering
      \includegraphics[width=1\textwidth]{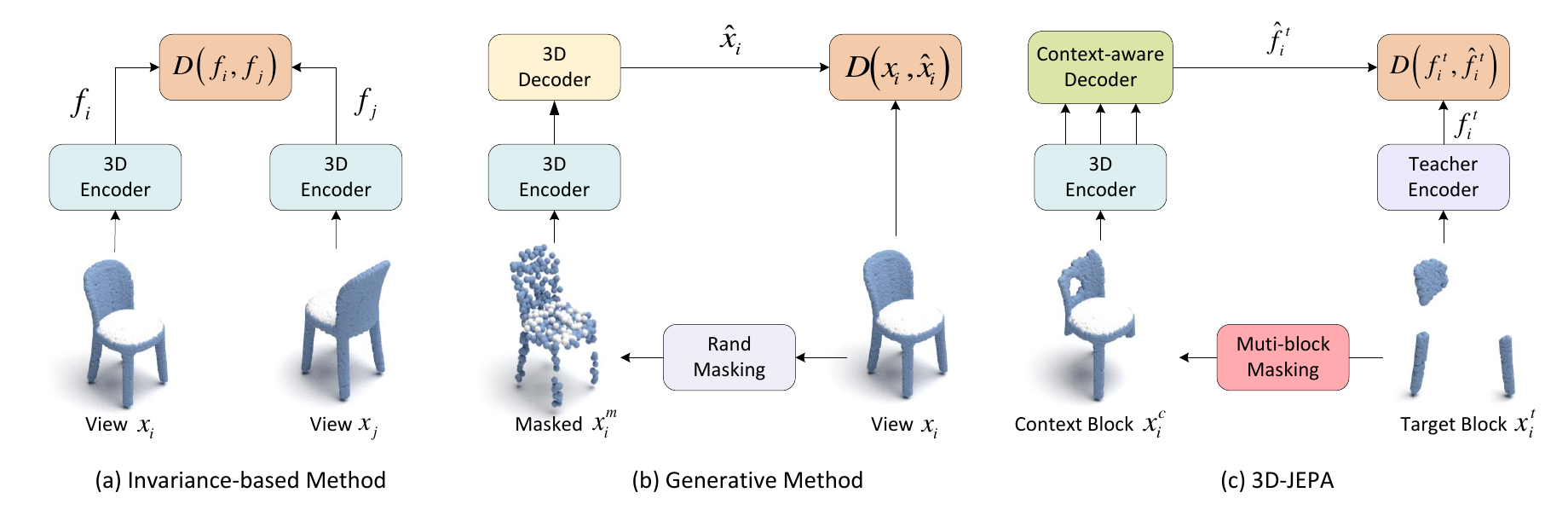}
      \vspace{-0.5cm}
      \caption{\textbf{Concept comparison of Invariance-based, Generative methods and our 3D-JEPA paradigms.} (a) Invariance-based methods aim to generate similar embeddings ${{f}_{i}}$ and ${{f}_{j}}$ for compatible input pairs ${{x}_{i}}$ and ${{x}_{j}}$. (b) After generating the masked point cloud $x_{i}^{m}$ by random masking, Generative methods aim to output embeddings ${{\hat{x}}_{i}}$ to predict the original data ${{x}_{i}}$ as much as possible. (c) After generating the target block $x_{i}^{t}$ and context block $x_{i}^{c}$ by muti-block masking, 3D-JEPA aims to output the embeddings $\hat{f}_{i}^{c}$ to predict the embeddings $f_{i}^{t}$ of $x_{i}^{t}$.  } 
    \label{3method}
      \vspace{-0.2cm}
   \end{figure*}

In cognition theory, humans learn an enormous amount of background knowledge about the world simply by passively observing it~\cite{lecun2022path}. For example, in the real world, humans cannot fully observe every part of a 3D object. A child who is learning can infer the type of an object simply by looking at part of it, while the best existing 3D vision models require thousands of complete training data. Even so, they fall short of human’s ability to perceive the world. Drawing inspiration from this phenomenon, ~\cite{assran2023self} proposed a non-generative approach for SSRL in 2D field. In the joint-embedding predictive architecture, the context encoder predicts the representation of various target blocks from a single context block in the same sample.

In this paper, we propose the first non-generative pretraining architecture of point cloud representation learning, 3D-JEPA. Firstly, to overcome the disadvantage of previous invariance-based and generative methods, we propose the multi-block sampling strategy to obtain a single context block and several target blocks with rich semantics in the same point cloud circumventing bias introduced by hand-crafted data augmentations. Then, the encoder predicts the high-level concepts of various target blocks from the context block in the feature space. Thereby avoiding generative methods that focus too much on irrelevant details (regular pattern). Secondly, point cloud have highly structured information (airplanes are always similar in shape). In generative architecture, the representation of the visible patches and the priori position information is only fed to the first layer of the decoder, resulting in the encoder having to capture the target-specific information of the visible patches. To solve this problem, we introduce the context-aware decoder which provides the context information to each layer of the decoder. Thus the encoder can focus on high-level semantic modeling that benefits for pre-training.

To summarize, the contributions of our paper include:

\begin{itemize}
    \item We propose 3D-JEPA, a novel 3D non-generative SSRL architecture. 3D-JEPA predicts the target blocks from the context block which extracts a high level of semantic representation during pretraining.
    
    \item We present the context-aware decoder that incorporates the context information in the decoder continuously which facilitates the encoder to model semantic information rather than the specific position information.
    
    \item Extensive experiments on various 3D downstream tasks demonstrate the effectiveness of  3D-JEPA. Our method uses half of the training epochs compared to previous methods but achieves superior performance.
\end{itemize}

\section{RELATED WORK}

\subsection{Invariance-based Method In 3D Vision}

As the Fig.~\ref{3method} (a) shows, the invariance-based methods optimize the encoder by outputting similar embeddings for compatible inputs, vice versa. Inspired by the self-supervised pretraining via contrastive learning in 2D vision~\cite{asano2019self,zbontar2021barlow,fu2022distillation}.  PointContrast~\cite{xie2020pointcontrast} is the pioneering method in 3D vision, allowing the network to learn equivalence to geometric transformations by contrasting points between two transformed views. Crosspoint~\cite{afham2022crosspoint} learns transferable 3D point cloud representation between the point cloud and its corresponding image based on contrastive learning. Pointclip~\cite{zhang2022pointclip} conducts alignment between CLIP-encoded point cloud and 3D category texts. The previously mentioned methods are remarkable, but they might introduce biases unsuitable for diverse downstream tasks, especially those involving different data distributions such as cropping and cutout during pretraining~\cite{assran2022hidden}. In our work, we do not set the hand-crafted data augmentations but seek to predict the representation of other parts in the same point cloud.

\subsection{Generative Method In 3D Vision}
   
\begin{figure*}
    \centering
     \includegraphics[width=\textwidth]{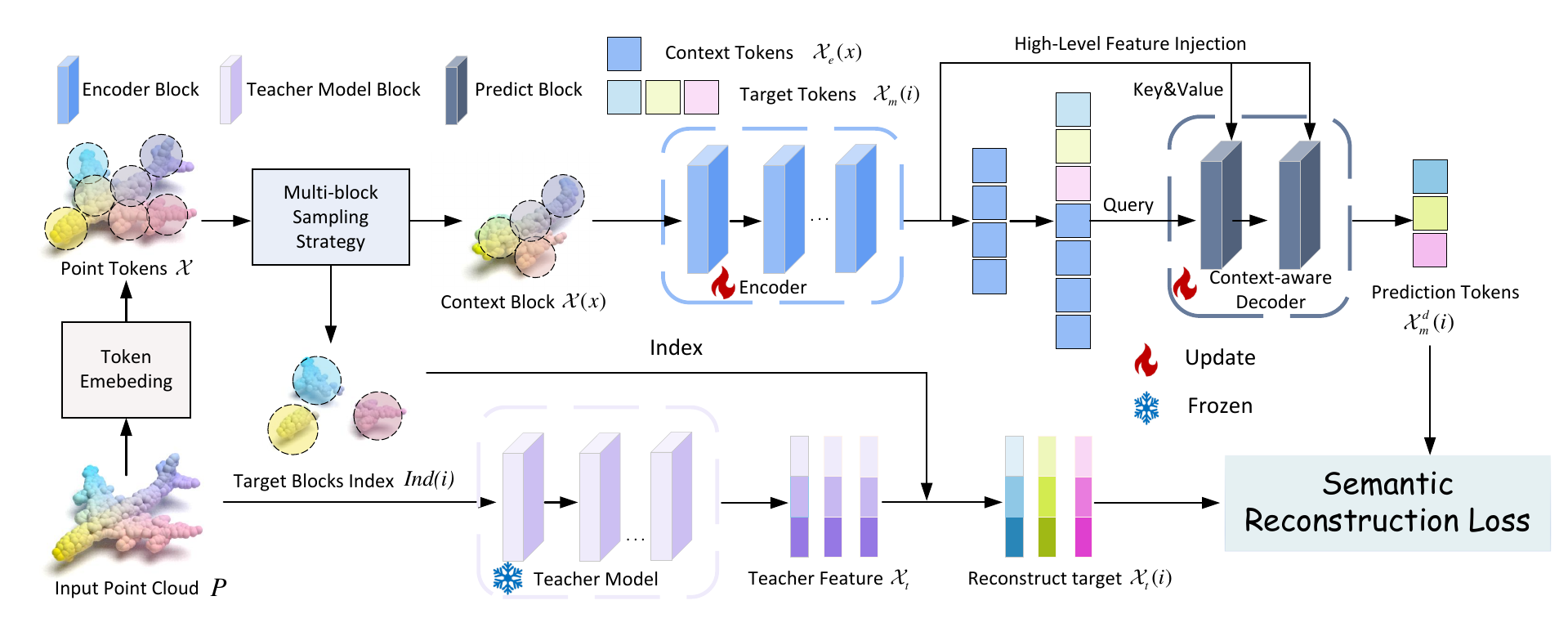}
     \vspace{-0.5cm}
    \caption{\textbf{The Pipeline of 3D-JEPA.} Given the input point cloud, the context block will be encoded as sequential tokens after the multi-block sampling. The context representation are then fed to every layer of the decoder to predict the representation of target blocks supervised by the outputs of the teacher model via cosine loss.}
    \label{pipeline}
    \vspace{-0.2cm}
\end{figure*}

Recently, motivated by the success of BERT~\cite{devlin2018bert} and MAE~\cite{he2022masked} in NLP and 2D vision, generative methods in 3D vision~\cite{qi2017pointnet,ma2022rethinking} have conquered the limited data domains problem in supervised learning. As shown in Fig.~\ref{3method} (b), most generative methods combine an encoder and a decoder, utilizing a standard Transformer to process visible point cloud patches and reconstruct missing input patches~\cite{min2023occupancy}. Point-MAE~\cite{pang2022masked} is an initiative along this direction that predicts the masked patches on the point level. Point-M2AE~\cite{zhang2022point} adopts a pyramid encoder and decoder architecture that can produce more hierarchically structured embeddings. ACT~\cite{dong2023autoencoders} leverages the self-supervised 3D transformers pretrained with 2D images to help 3D representation learning through knowledge distillation. Occ-BEV~\cite{min2024multi} reconstructs the 3D scene as the foundational stage and subsequently finetunes the model on downstream tasks. In contrast to those approaches, our architecture predicts the global representation of target blocks instead of focusing on predicting every masking token, thereby avoiding attention to unnecessary details.

\subsection{Joint Embedding Predictive Architecture}

The core idea of Joint Embedding Predictive Architectures (JEPA) is recently provided by ~\cite{lecun2022path}, which is similar to invariance-based and generative methods. As shown in Fig.~\ref{3method} (c), the key difference is that the reconstruction objective of JEPA is the abstract semantic representation rather than raw data. I-JEPA~\cite{assran2023self} is first proposed to predict the representation of various target blocks from a context block in the same image. MC-JEPA~\cite{bardes2023mc} is a multi-task approach to jointly learn optical flow and content features. To explore the 3D application of the JEPA, we design a multi-block sampling strategy that can sample semantic blocks from the point cloud and predict the representation of target blocks in embedding space. Point2Vec~\cite{zeid2023point2vec} predicts the representation of random missing point patches through an online target encoder. Compared with Point2Vec, the prediction target obtained by the multi-block sampling strategy contains richer context semantic information which exhibits significant improvements in effectiveness and efficiency.

\section{METHOD}

An overview of our 3D-JEPA framework is illustrated in Fig.~\ref{pipeline}. In Section~\ref{embed}, we first introduce the token embedding module processing the input point cloud. Then, Section~\ref{sample} presents the multi-block sampling strategy of 3D-JEPA that obtains a context block and multi target blocks in the same point cloud. Then, in Section~\ref{architecture}, it shows the details of the encoder and context-aware decoder. Finally, Section~\ref{loss} describes the loss function.

\subsection{Token Embedding}
\label{embed}
Due to the unorder of point cloud and quadratic complexity of self-attention operators, we adopt the patch embedding strategy~\cite{yu2022point} that converts input point cloud into 3D point patches instead of inputting all point cloud into the Encoder directly. Given a raw point cloud $P \in \mathbb{R}^{N \times 3}$ with $N$ points encoded in $(x, y, z)$ Cartesian space, we first sample $M$ center points $CT \in \mathbb{R}^{M \times 3}$ using farthest point sampling (FPS). Then, we utilize K Nearest-Neighbour (K-NN) to gather the $K$ nearest neighbors for each center point, dividing the point cloud into the corresponding point patches $\mathcal{N}=\left\{\mathcal{N}_{i} \mid i=1,2, \ldots, M\right\} \in \mathbb{R}^{M \times \mathbb{K} \times 3}$. We further aggregate the point patches $\mathcal{N}$ by a lightweight PointNet~\cite{qi2017pointnet} to obtain point tokens $\mathcal{X}=\left\{\mathcal{X}_{i} \mid i=1,2, \ldots, M\right\} \in \mathbb{R}^{M \times \mathbb{C}}$ where $\mathcal{X}_{i}$ is the representation associated with the patch $\mathcal{N}_{i}$ and the $C$ is feature dimension. We will feed the point tokens $\mathcal{X}$ to the following encoder and teacher model.

\subsection{Multi-block Sampling Strategy}
\label{sample}

\begin{figure}[t]
\vspace{0.1cm}
\centering
    \includegraphics[width=0.95\linewidth]{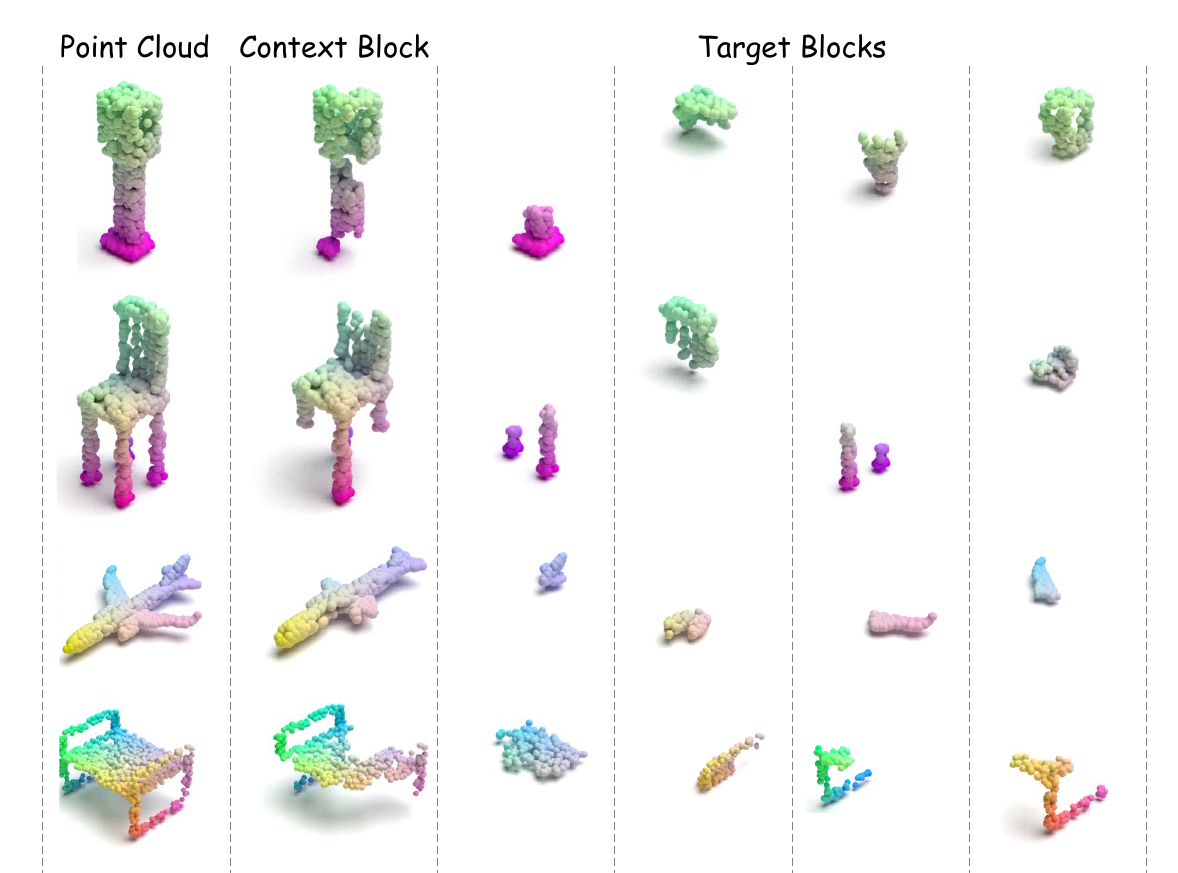}
  \caption{\textbf{Visualization of multi-block sampling.} Given the point cloud, we sample 4 target blocks via FPS with a lower scale. Next, we randomly sample a context block with a larger scale and remove any overlapping target blocks. In this way, the target blocks have global semantic information, and the context block is informative.}
  \label{visio}
  \vspace{-0.3cm}
\end{figure}

Generative methods commonly randomly mask the point tokens at a large ratio e.g., 75\%-80\%. In contrast to the aforementioned approaches, we separately sample the target blocks and context block using the multi-block sampling strategy from the same point cloud. We visualize the multi-block sampling strategy of 3D objects in Fig.~\ref{visio}.

\subsubsection{Target Blocks Sampling}
We first describe how we produce the target blocks in the 3D-JEPA framework. Compared to random sampling, we expect to obtain target blocks with a lower overlapping rate, thus avoiding predicting semantically similar representation thereby improving efficiency. We sample $A$ points via FPS as target blocks center from $CT$ and select the nearest tokens in the range $(0.15,0.2)$ via K-NN. By the methods above, we obtain the target blocks index $Ind(i)=\left\{CT_{j}\right\}_{j \in B_{i}}$, where the $B_{i}$ denote the sampling corresponding of the $i^\text{th}$ target block.

\subsubsection{Context Block Sampling}
To ensure obtaining a sufficiently informative context block, we sample a single block $\mathcal{X}(x)=\left\{\mathcal{X}_{j}\right\}_{j \in B_{x}}$ from the point token $\mathcal{X}$ with a large scale in the range $(0.85, 1.0)$, where the $B_{x}$ associated with the context block. Since we sample the target and context blocks independently, this leads to the leakage of target information which makes the predicting task less challenging. Thus, we remove any overlapping tokens from the context block. Meanwhile, it reduces the encoder high consumption of computing resources. 

\subsection{Model Architecture}
\label{architecture}

Similar to most generative models~\cite{he2022masked,pang2022masked}, the 3D-JEPA consists of an encoder and Context-aware decoder.
\subsubsection{Encoder}
The encoder aims to comprehend the global spatial geometries with rich semantic representation that consists of Standard Transformers~\cite{vaswani2017attention} with self-attention layers. After the context block $\mathcal{X}(x)$ is added to the corresponding positional embedding $POS(x)$, it is further fed to the encoder mapping to the corresponding representation:
\begin{equation}
\begin{split}
    \mathcal{X}_{e}(x)=\operatorname{Encoder}(\mathcal{X}(x), POS(x)).
\end{split}
\end{equation}

\subsubsection{Context-aware Decoder}
\begin{figure}[t]
\vspace{-0.3cm}
\centering
    \includegraphics[width=0.49\textwidth]{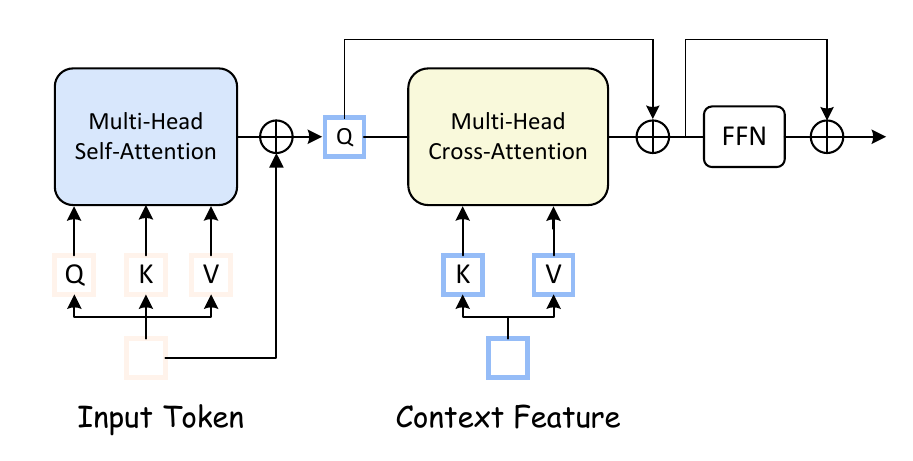}
  \caption{\textbf{Context-aware Block.} In the decoder block, we implement cross-attention layers after the self-attention. }
  \label{transformer}
  \vspace{-0.3cm}
\end{figure}

Previous generative work concatenates or adds the encoded $\mathcal{X}_{e}(x)$ with a set of shared learnable tokens $\mathcal{X}_{m}(i)$. Then jointly feed them to the first layer of the decoder. We argue that in this way, the $\mathcal{X}_{e}(x)$ is invisible to deeper layers in the decoder during feature prediction. Resulting in the encoder considers memorizing the context information of the context blocks, which limits the encoder modeling capability to learn structure knowledge. The Context-aware decoder is designed to feed the context representation $\mathcal{X}_{e}(x)$ to each layer of the decoder with a cross-attention mechanism after self-attention in every decoder block. As illustrated in Fig.~\ref{transformer}, we first input Multi-Head Self-Attention both encoded tokens and target tokens added positional embeddings, the output of Multi-Head Self-Attention is treated as the query array. While context representation $\mathcal{X}_{e}(x)$ the is treated as the key array and value array of cross-attention. The context-aware decoder structure is formulated as,
 
\label{encoder}

\begin{equation}
\label{eq:dec}
\begin{split}
    (\mathcal{X}^d_{e}(x),\mathcal{X}^d_{m}(i))=\operatorname{Decoder}\left(\mathcal{X}_{e}(x), \mathcal{X}_{m}(i)\right)).
\end{split}
\end{equation}

We repeat the Eq.~\ref{eq:dec} $A$ times to generate the predictions $\mathcal{X}^d_{m}(1),...,\mathcal{X}^d_{m}(A)$, aiming to reconstruct corresponding the semantic representation of the target blocks.

\subsection{Objective Function}
\label{loss}

\subsubsection{Reconstruction Target}To ensure that the target representation has global information about the point cloud but just in a local pattern, we feed all the point tokens $\mathcal{X}$ in the teacher model $f_{\mathcal{T}}$ to obtain the corresponding representation $\mathcal{X}_{t}=\left\{\mathcal{X}_{i} \mid i=1,2, \ldots, M\right\} \in \mathbb{R}^{M \times \mathbb{C}_{t}}$, where the $\mathbb{C}_{t}$ is the dimension of the teacher model. To obtain the reconstructed target $\mathcal{X}_{t}(i)$, we utilize the index of the target blocks $Ind(i)$ to aggregate the corresponding output of the teacher model by channel-wise concatenation. A single linear projection layer $FC(\cdot)$ is adopted to map the output of the decoder $\mathcal{X}^d_{m}(i)$ and reconstruct target $\mathcal{X}_{t}(i)$ to the same feature space. 
\subsubsection{Loss Function}The knowledge distillation loss minimizes the negative cosine similarity between the prediction and target features: 
\begin{equation}
\begin{split}
    \mathcal{L}_{r e c}=-\frac{1}{A}\sum_{i=1}^{A} \mathcal{L}_{c o s}\left(FC(\mathcal{X}^d_{m}(i)), FC(\mathcal{X}_{t}(i))\right), 
\end{split}
\end{equation}
where the $\mathcal{L}_{\cos }(s, t)=1-\frac{s \cdot t}{|s||t|}$.

With such constraints, 3D-JEPA not only explores the semantic knowledge but also ignores the unnecessary detailed representation thus improving efficiency.

\section{Experiments}

\subsection{Pretraining Setting}

For a fair comparison with previous work~\cite{qi2017pointnet,yu2022point}, we pretrain our 3D-JEPA on the ShapeNet~\cite{chang2015shapenet}. The dataset comprises more than 50,000 CAD models from 55 object categories. We resample the source point cloud to 2,048 points that only contain $(x,y,z)$ coordinate information via FPS and use scale and rotation to augment the input data. We apply the encoder with a 12-layer Standard Transformer block and the context-aware decoder with a 2-layer block.

In line with ACT~\cite{dong2023autoencoders} which transfers the pretrained foundational Transformers as cross-model 3D teacher, we adopt the the dVAE tokens from the tuned 3D autoencoder as the reconstructed target.

We use the AdamW optimizer with a learning rate value of 0.001. All experiments are performed on a NVIDIA 3090 GPU. After pretraining, we employ the encoder for downstream tasks.

\subsection{Downstream Tasks}
\begin{table}[h]
\caption{\textbf{Classification results on the ScanObjectNN~\cite{yi2016scalable} and ModelNet40\cite{wu20153d}.} \#EP denotes the epochs of inference model during pre-training.  }
\setlength{\abovecaptionskip}{0pt}
\setlength{\belowcaptionskip}{0pt}
\vspace{-0.1cm}
\begin{center}
        {   \resizebox{1\linewidth}{!}{
                \begin{tabular}{lcccccc}
\toprule[0.95pt]
\multirow{2}{*}[-0.5ex]{Method} & \multirow{2}{*}[-0.5ex]{\#EP} &     \multicolumn{3}{c}{ScanObjectNN} & \multicolumn{1}{c}{ModelNet40}\\
\cmidrule(lr){3-5}\cmidrule(lr){6-6} &  & OBJ\_BG & OBJ\_ONLY & PB\_T50\_RS& 1k P\\
\midrule[0.6pt]
\multicolumn{6}{c}{\textit{Supervised Learning Only}}\\
\midrule[0.6pt]
PointNet~\cite{qi2017pointnet} & -  & 73.3 & 79.2 & 68.0 & 89.2\\ PointNet++~\cite{qi2017pointnet++} & - & 82.3 & 84.3 & 77.9 & 90.7\\
DGCNN~\cite{wang2019dynamic} & -  & 82.8 & 86.2 & 78.1 & 92.9\\
PointNeXt~\cite{qian2022pointnext} & -  & - & - & 87.7$\pm$0.4 & 94.0 \\


\midrule[0.6pt]
\multicolumn{6}{c}{\textit{with Self-Supervised Representation Learning} ({\scshape Full})}\\

\midrule[0.6pt]

Transformer~\cite{vaswani2017attention} & 300  &  83.04 & 84.06 & 79.11 & 91.4\\
Point-BERT~\cite{yu2022point} & 300  & 87.43 & 88.12 & 83.07 & 93.2\\
Point-MAE~\cite{pang2022masked} & 300 & 90.02 & 88.29 & 85.18 & 93.8\\
Point-M2AE~\cite{zhang2022point} & 300  & 91.22 & 88.81 & 86.43 & 94.0\\
Point2Vec~\cite{zeid2023point2vec} & 300  & 91.2 & 90.4 & 87.5 & \textbf{94.8} \\
ACT~\cite{dong2023autoencoders} & 300  &  93.29 & 91.91 & 88.21 & 93.7\\
ViPFormer~\cite{10160658} & 300  &  90.7 & - & - & 93.9\\
 
3D-OAE ~\cite{10610588} & 300  & 89.16 & 88.64 & 83.17 & 93.4\\

\rowcolor{blue!5} \textbf{3D-JEPA}  & 150  & 93.80 & 92.77 & 88.65 & 93.8\\

\rowcolor{blue!5}\textbf{3D-JEPA}  & 300  & \textbf{94.49} & \textbf{93.63} & \textbf{89.52} & 94.0\\
                    \bottomrule
                    \end{tabular}}
        }
\end{center}
\label{tab:scanobjectnn}
\vspace{-0.3cm}
\vskip -5pt
\end{table}

\subsubsection{Transfer Protocol}
 We connect the classification head consisting of a 3-layer non-linear MLP after the pretrained model and update all the parameters of the encoder and the classification head.

\subsubsection{3D Real-World Object Classification}
\begin{table}
    \centering
    \caption{\textbf{Few-shot Classification with Standard Transformers on ModelNet40 dataset.} }
    \label{tab:fewshot}
    \setlength\tabcolsep{2pt}
    \resizebox{\linewidth}{!}{
    \begin{tabular}{lcccc}
    \toprule[0.95pt]
    \multirow{2}{*}[-0.5ex]{Method}& \multicolumn{2}{c}{5-way} & \multicolumn{2}{c}{10-way}\\
    \cmidrule(lr){2-3}\cmidrule(lr){4-5} & 10-shot & 20-shot & 10-shot & 20-shot\\
    \midrule[0.6pt]
     Transformer~\cite{vaswani2017attention} & 87.8 $\pm$ 5.2& 93.3 $\pm$ 4.3 & 84.6 $\pm$ 5.5 & 89.4 $\pm$ 6.3\\
     Point-BERT~\cite{yu2022point} & 94.6 $\pm$ 3.1 & 96.3 $\pm$ 2.7 &  91.0 $\pm$ 5.4 & 92.7 $\pm$ 5.1\\
     Point-MAE~\cite{pang2022masked} & 96.3 $\pm$ 2.5 & 97.8 $\pm$ 1.8 & 92.6 $\pm$ 4.1 & 95.0 $\pm$ 3.0\\
     Point-M2AE~\cite{zhang2022point}  &96.8\ $\pm$\ 1.8 &98.3\ $\pm$\ 1.4 &92.3\ $\pm$\ 4.5 &95.0\ $\pm$\ 3.0\\
     Point2Vec~\cite{zeid2023point2vec} &97.0\ $\pm$\ 2.8 &98.7\ $\pm$\ 1.2 &93.9\ $\pm$\ 4.1 &95.8\ $\pm$\ 3.1\\
    ACT~\cite{dong2023autoencoders}  &96.8\ $\pm$\ 2.3 &98.0\ $\pm$\ 1.4 &93.3\ $\pm$\ 4.0 &95.6\ $\pm$\ 2.8\\
    
    3D-OAE ~\cite{10610588}  &96.3\ $\pm$\ 2.5 &98.2\ $\pm$\ 1.5 &92.0\ $\pm$\ 5.3 &94.6\ $\pm$\ 3.6\\
     \rowcolor{blue!5} \textbf{3D-JEPA(150 Epoch)}  &\textbf{97.6\ $\pm$\ 2.0}& \textbf{98.8\ $\pm$\ 0.4}&\textbf{94.3\ $\pm$\ 3.6} &\textbf{96.3\ $\pm$\ 2.4}\vspace{0.1cm}\\
     \bottomrule[0.95pt]
     \end{tabular}
    }
    \vspace{-0.3cm}
\end{table}

We finetune the encoder for classification tasks and report the overall accuracy without the voting strategy on Real-world datasets: ScanObjectNN~\cite{yi2016scalable}. ScanObjectNN is a challenging 3D dataset consisting of 11,416 training and 2,882 test 3D shapes, which includes backgrounds with noise. We conduct experiments on three splits of ScanObjectNN, namely OBJ-BG, OBJ-ONLY, and PB-T50-RS. It can be observed from TABLE~\ref{tab:scanobjectnn} that: (1) Comparing the Transformer baseline~\cite{vaswani2017attention}, the 3D-JEPA achieves a significant improvement of +$31.43\%$ accuracy on the three variant ScanObjectNN benchmarks. (2) Compared to previous SSRL methods ~\cite{pang2022masked}, ~\cite{zeid2023point2vec}, which use 300 epochs during pretraining. Our method just uses 150 epochs which can produce enhancements efficiently. (3) The 3D-JEPA only leverages the single-modal information achieving the best generalization compared to other cross-modal SSRL methods, e.g. ViPFormer~\cite{10160658} is pretrained by optimizing intra-modal and cross-modal contrastive objectives. (4) Compared to methods adopt pyramid encoder and decoder architecture, such as point-M2AE~\cite{zhang2022point}, 3D-JEPA achieves improvement while remaining efficient (a 3× increase in processing speed).

\begin{figure}[t]
\vspace{0.1cm}
\centering
    \includegraphics[width=0.99\linewidth]{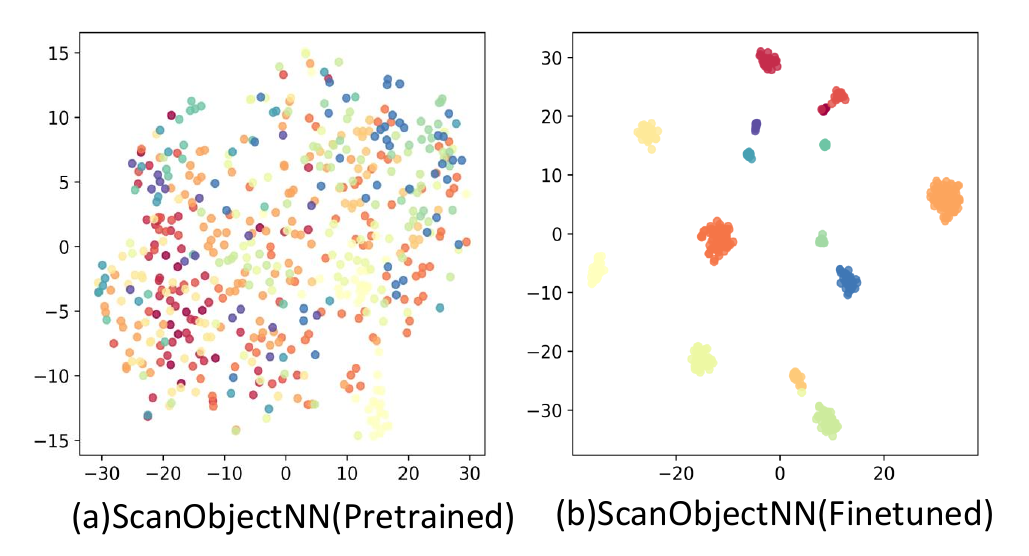}
    \vspace{-0.3cm}
  \caption{\textbf{t-SNE feature for ScanObjectNN BJ datasets and ablation study results for Context-aware decoder.}}
  \vspace{-0.2cm}
  \label{tsne}
\end{figure}

As illustrated in Fig.~\ref{tsne} (a) and (b), different colors indicate different classes. Feature vectors extracted by fine-tuning model are clustered according to the labels. This means that we can extract high-dimensional semantic information of point cloud in downstream tasks.

\subsubsection{3D Synthetic Object Recognition}

We construct the experiment on ModelNet40 \cite{wu20153d} to evaluate the understanding ability of synthetic datasets. ModelNet40 is obtained by sampling 3D CAD models, and it contains 12,331 objects (9,843 for training and 2,468 for testing) from 40 categories. We use scale and translation as data augmentations. As TABLE~\ref{tab:scanobjectnn}, we can obtain the accuracy close to the state of the art in previous SSRL. 

We further conduct experiments for few-shot classification on ModelNet40 using only few available labels. Following the common routine~\cite{yu2022point}, we randomly selected N classes from the dataset and selected M samples in each class. TABLE~\ref{tab:fewshot} shows the results that reported the mean and standard deviation over 10 runs. We can see (1) Our method brings significant improvements of +9.8\%, +5.5\%, +9.7\%, and +6.9\% over the Transformer baseline~\cite{vaswani2017attention}. (2) 3D-JEPA outperforms previous SSRL methods while requiring fewer pretraining epochs on all settings. 
\begin{table}
\caption{\textbf{Part Segmentation on ShapeNetPart \protect\cite{yi2016scalable}}. }
\vspace{0.1cm}
\small
\centering
\centering
	\begin{tabular}{lccc}
	\toprule
		Method   &mIoU$_C$ &mIoU$_I$\\
		\cmidrule(lr){1-1} \cmidrule(lr){2-2} \cmidrule(lr){3-3}  
	    PointNet~\cite{qi2017pointnet}  &80.39 &83.70 \\
	    PointNet++~\cite{qi2017pointnet++} &81.85 &85.10 \\
	    DGCNN~\cite{wang2019dynamic}  &82.33 &85.20 \\
	    \cmidrule(lr){1-3}
        
	    Transformer~\cite{vaswani2017attention}  &83.42 &85.10 \\
	    Point-BERT~\cite{yu2022point}  &84.11 &85.60 \\
	    Point-MAE~\cite{pang2022masked}  &- &86.10\\
        Point2Vec~\cite{zeid2023point2vec}  &84.6 &86.3\\
        Point-M2AE~\cite{zhang2022point}  &84.86 &86.51\\
        ACT~\cite{dong2023autoencoders}  &84.66 &86.14\\
        ViPFormer~\cite{10160658}  &- &84.7\\
        3D-OAE ~\cite{10610588} &- &85.7\\
	    \rowcolor{blue!5} \textbf{3D-JEPA(150 Epoch)}  &84.73 &86.28\\
     \rowcolor{blue!5} \textbf{3D-JEPA(300 Epoch)} &\textbf{84.93} &\textbf{86.41}\vspace{0.1cm}\\
	  \bottomrule
	\end{tabular}

\label{tab:seg}
\vspace{-0.3cm}
\end{table}

\subsubsection{Part Segmentation}
\begin{figure*}[t!]
  \centering
    \includegraphics[width=\textwidth]{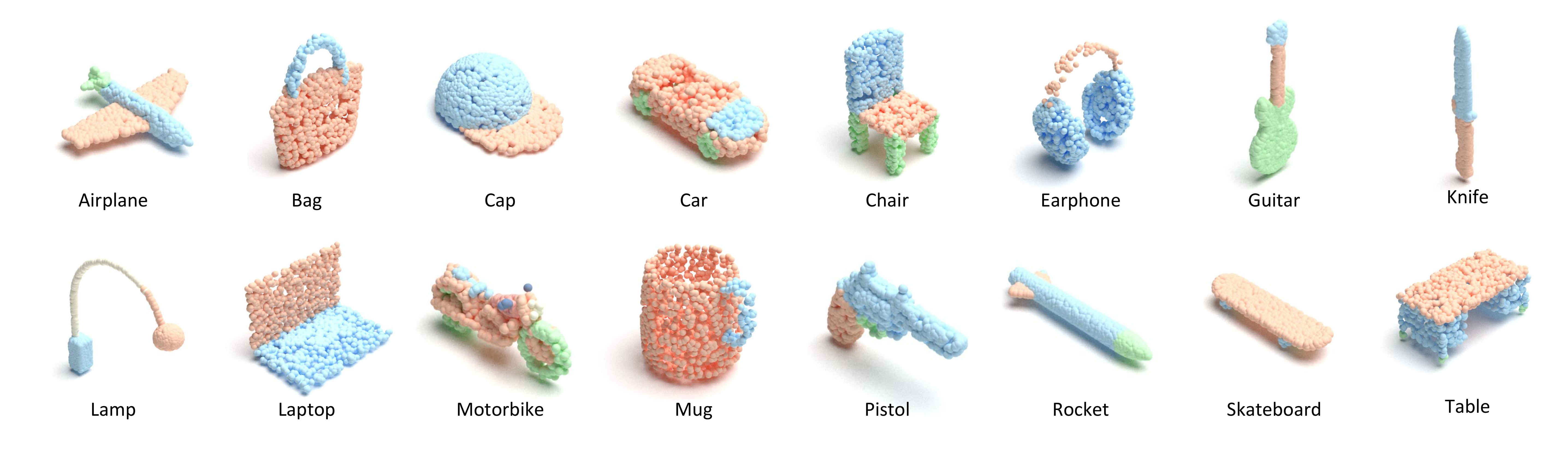}
    
    \caption{\textbf{Illustration of part segmentation results.}}
    \label{seg}
    
\end{figure*}

Compared with the classification tasks, part segmentation tasks are more challenging. To evaluate the scene geometry semantic understanding performance within 3D objects of 3D-JEPA, we conduct 3D part segmentation on ShapeNetPart~\cite{chang2015shapenet}, which contains 16,881 instances of 16 categories. We report the mean IoU across all part categories (mIoU$_C$) and all instances (mIoU$_I$) respectively in the TABLE~\ref{tab:seg}. It can be observed that 3D-JEPA improves the Transformer baseline by +1.51\% (mIoU$_C$) and +1.31\% (mIoU$_I$). It shows that predicting high-level semantic representation of the target blocks is still efficient and handy in the part segmentation task. In addition, Fig.~\ref{seg} visualizes the part segmentation of ShapeNetPart. 

\subsection{Ablation Study}

\begin{table}[t!]
\caption{\textbf{Ablation Results on Sampling strategy}.}

\centering
\small
	\begin{tabular}{ccccc}
	\toprule
 \multicolumn{2}{c}{\ \ \ \ \ \ \ \ \ \ \ \ \ \ \ \ Sampling Strategy\ \ \ \ \ \ \ \ \ \ \ \ \ \ \ \ } &\makecell*[c]{\multirow{2}*{\ \ \ Acc. (\%)\ \ \ }} \\
 \cmidrule(lr){1-2} 
    \ \ \ \ \ \ \ Sampling Method\ \ \ \ \ \ \  &Target Num & \\
		 \cmidrule(lr){1-1}  \cmidrule(lr){2-2} \cmidrule(lr){3-3}
            Rand&- & 91.91\\
            Block &-  &91.57\\
            Muti-blocks &3 &93.35\\
            \rowcolor{blue!5} Muti-blocks &4 &\textbf{93.63}\\

	  \bottomrule
	\end{tabular}

\label{ab}
\end{table}

In this section, we study the impact of each major component in 3D-JEPA. We report the results of several ablation experiments on the OBJ-ONLY benchmark.

\subsubsection{Multi-block Sampling Strategy}

TABLE \ref{ab} shows the impact of the multi-block sampling strategy in 3D-JEPA, which is compared with the random masking and block masking typically used in generative method ~\cite{pang2022masked}. In random masking, the target is a set of random point patches and the context is the point complement. In block masking, the target is randomly sampled multi neighboring point patches and the context is point complement. Among them, block masking is similar to 3D-JEPA, with the difference being that 3D-JEPA only requires processing a single part of each point cloud. This sampling approach is more consistent with the human cognition of inferring the kind of 3D object from a single view. We set the masking ratio 0.25 in random and block masking strategy. In TABLE \ref{ab}, we can see that our sampling strategy predicts representation of multi blocks can help the encoder learn semantic representation.

\begin{table}[t!]
\caption{
\textbf{Ablation Results on Decoder}.
}
\vspace{-0.15cm}
\centering
\small
	\begin{tabular}{ccccc}
	\toprule
 \multicolumn{2}{c}{\ \ \ \ \ \ \ \ \ \ \ \ \ \ \ \ Decoder\ \ \ \ \ \ \ \ \ \ \ \ \ \ \ \ } &\makecell*[c]{\multirow{2}*{\ \ \ Acc. (\%)\ \ \ }} \\
 \cmidrule(lr){1-2} 
    \ \ \ \ \ \ \ Context-aware\ \ \ \ \ \ \  & Decoder Depth & \\
		 \cmidrule(lr){1-1}  \cmidrule(lr){2-2} \cmidrule(lr){3-3}
            -&2 & 93.12\\
            \checkmark & 0  &92.77\\
            \rowcolor{blue!5} \checkmark & 2 &\textbf{93.63}\\
            \checkmark & 4 &93.29\\
	  \bottomrule
	\end{tabular}

\label{ab1}
\end{table}


We also discussed the number of target blocks in the multi-block sampling strategy. We consider that when the number is less, it is impossible to predict the full semantic information of the point cloud. In contrast, when the number is further increased, it will introduce additional tasks making the model complex.

\subsubsection{Decoder}

As TABLE~\ref{ab1} shows, the model with the Context-aware decoder achieves better accuracy. It can be demonstrated that this module helps the encoder reconstruct the semantic representation of target blocks.

We also examine the impact of the decoder depth on the pretraining stage. TABLE~\ref{ab1} shows that the depth of the decoder does not have a significant impact on the encoder's ability and when the decoder depth is set to '2' getting the best results. It's worth noting that when the decoder depth is '0', we put the context tokens and target tokens together in the encoder leading to an inferior result.



\section{CONCLUSIONS}

In this paper, we propose the first non-generative framework for 3D SSRL. The 3D-JEPA circumvents bias introduced by hand-crafted data augmentations and focuses on necessary high-level semantic information. Firstly, the multi-block sampling strategy effectively extracts context and target blocks from the same point cloud. Secondly, the Context-aware Decoder aids the encoder in better capturing of high-level semantic representation. The experiments demonstrate 3D-JEPA's outstanding performance across various downstream tasks with less latency. 


\bibliographystyle{IEEEtran}
\bibliography{IEEEabrv}

\end{document}